\documentclass[letterpaper, 10 pt, conference]{ieeeconf}  

\IEEEoverridecommandlockouts                              

\usepackage{placeins}
\usepackage{graphicx}
\usepackage{amsmath,amssymb,amsfonts}
\usepackage{algpseudocode}
\usepackage{float}
\usepackage{algorithm}
\usepackage{graphicx}
\usepackage{textcomp}
\usepackage{xcolor}
\usepackage{hyperref}
\usepackage{subfigure}
\usepackage{caption}
\usepackage{booktabs}
\usepackage{comment}
\usepackage{gensymb}
\usepackage{cleveref}
\usepackage{pgfplots}
\pgfplotsset{compat=1.17}

\definecolor{EyeInFingerColor}{RGB}{0, 102, 204} 
\definecolor{RealsenseColor}{RGB}{204, 51, 0}  

\crefname{section}{Sec.}{Section.}
\Crefname{section}{Sec.}{Section.}
\crefname{figure}{Fig.}{Figure.}
\Crefname{figure}{Fig.}{Figure.}
\crefname{table}{Table.}{Table.}
\Crefname{table}{Table.}{Table.}
\crefname{equation}{}{}
\Crefname{equation}{Equation}{Equations}

\newcommand{\ie}{\textit{i}.\textit{e}., }

\newcommand{\LEGO}{LEGO}

\title{\LARGE \bf
Eye-in-Finger: Smart Fingers for \\Delicate Assembly and Disassembly of \LEGO{}
}

\author{
Zhenran Tang,
Ruixuan Liu,
Changliu Liu
\thanks{This work is in part supported by the Manufacturing Futures Institute, Carnegie Mellon University, through a grant from the Richard King Mellon Foundation.}
\thanks{Zhenran Tang, Ruixuan Liu, and Changliu Liu are with Robotics Institute,
	Carnegie Mellon University,
	Pittsburgh, PA, USA.
        {\tt\footnotesize \{zhenrant,ruixuanl,cliu6\}@andrew.cmu.edu}}%
}

\begin{document}
\maketitle

\begin{abstract}

Manipulation and insertion of small and tight-toleranced objects in robotic assembly remain a critical challenge for vision-based robotics systems due to the required precision and cluttered environment. 
Conventional global or wrist-mounted cameras often suffer from occlusions when either assembling or disassembling from an existing structure.
To address the challenge, this paper introduces ``Eye-in-Finger", a novel tool design approach that enhances robotic manipulation by embedding low-cost, high-resolution perception directly at the tool tip. 
We validate our approach using \LEGO{} assembly and disassembly tasks, which require the robot to manipulate in a cluttered environment and achieve sub-millimeter accuracy and robust error correction due to the tight tolerances. 
Experimental results demonstrate that our proposed system enables real-time, fine corrections to alignment error, increasing the tolerance of calibration error from 0.4mm to up to 2.0mm for the \LEGO{} manipulation robot.
\end{abstract}

\section{Introduction}

\begin{figure*}
    \centering
    \includegraphics[width=\linewidth]{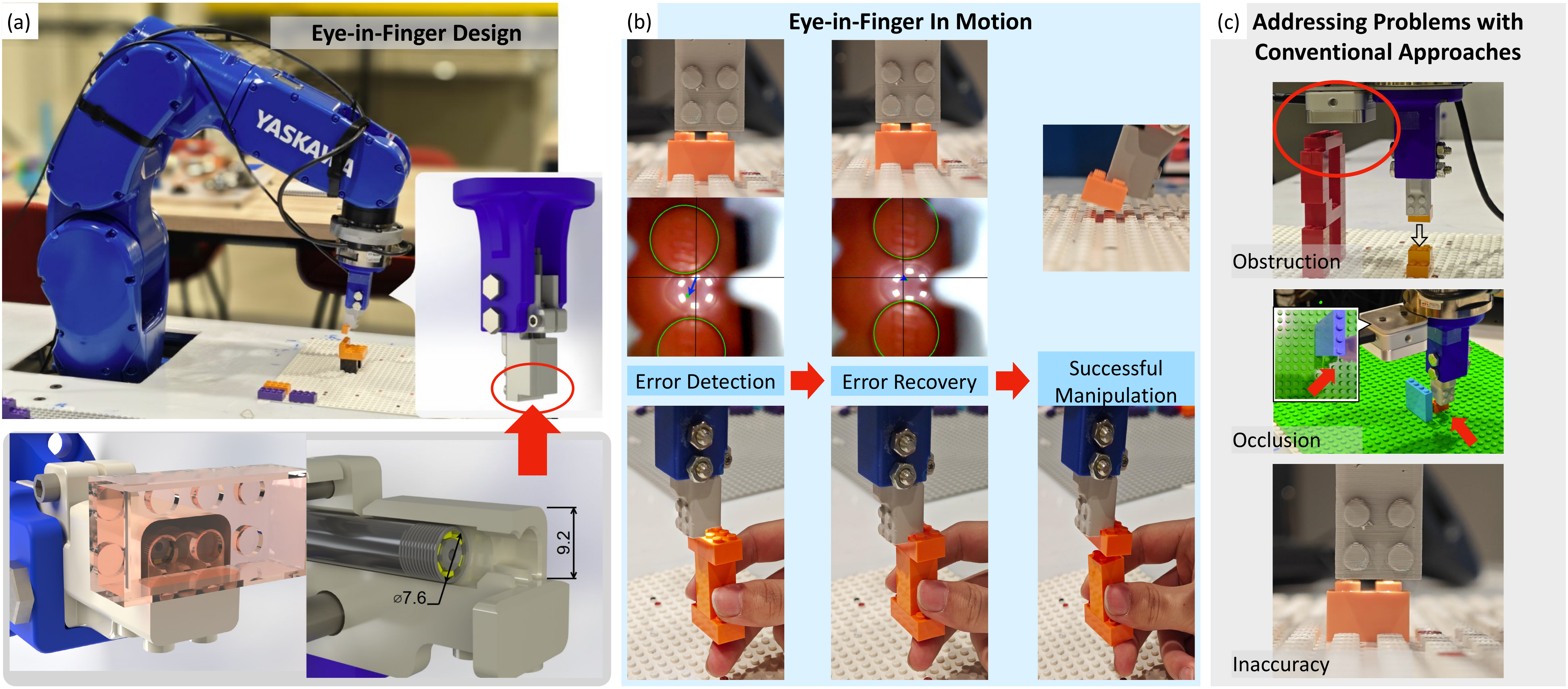}
    \caption{ \textbf{(a):} Top: EiF on a robot, holding a \LEGO{} brick. Bottom: Cut-out views of EiF with key dimensions in millimeters. \textbf{(b):} A robot using EiF to correct misalignment and pick up \LEGO{} bricks from a static board (top) and a moving hand (bottom). \textbf{(c):} Challenges with conventional approaches in \LEGO{} manipulation that EiF addresses.}
    \label{fig:overview-figure}
    \vspace{-15pt}
\end{figure*}

Humans rely on vision for overall spatial perception but depend on tactile sensing for fine-grained, high-precision interactions \cite{Yoshida2003}. For example, when threading a needle, placing a microchip on a circuit board, or performing delicate sutures in surgery, visual guidance provides an initial estimate, while tactile feedback refines positioning and detailed operations. In robotic systems, visual feedback similarly serves as a powerful tool for global perception, enabling high-level scene understanding and object localization\cite{fu2024mobile, black2024pi_0}. However, for precise manipulation of small objects, traditional vision-based approaches face critical challenges related to accuracy and occlusion.

Robotic manipulation of small objects presents multiple challenges, among which occlusion and precision are particularly important. In fine-detail tasks such as assembling with small components or performing surgery, the end effector is often larger than the target components, increasing the probability of occlusion when utilizing vision guidance. Moreover, vision systems often fail to provide the position estimation accuracy required for precise manipulation of small objects \cite{tang2024}. While tactile sensing offers high-resolution local perception, it suffers from inherent limitations, including high cost, durability concerns, and prolonged feedback latency \cite{Zhao2024}. Additionally, tactile sensors typically necessitate the use of soft materials at the point of contact, imposing constraints on mechanical design and further restricting their applicability in industrial and medical settings.

To bridge this gap, we introduce \textbf{Eye-in-Finger (EiF)}, a vision-integrated sensing method that delivers low-cost, high-resolution perception directly from the tip of the end effector.  Our approach provides precise spatial information in translation and rotation, achieving levels of accuracy traditionally reserved for contact-based tactile sensing. By embedding vision at the contact point, EiF effectively overcomes occlusion challenges and enables real-time, fine adjustments during delicate manipulation tasks. This approach allows robots to refine their actions dynamically, just as a human does by combining visual guidance with fingertip tactile feedback.

In this work, we use robotic \LEGO{} assembly and disassembly as a structured test case to evaluate the effectiveness of the EiF approach. \LEGO{}-based manipulation demands \textit{sub-millimeter} accuracy due to the tight tolerances of its interlocking mechanisms, making it an ideal benchmark for testing high-precision robotic systems. \cref{fig:overview-figure}(b) illustrates the precision constraint. If the tool is not precisely aligned with the brick, it will collide with the brick and likely damage the object. In addition, if bricks are not firmly pressed together, a minuscule tilt can occur as shown in \cref{fig:tilt1}, which needs to be detected promptly.  Popular setup with the use of wrist mounted camera are prone to both visual occlusion and physical obstruction, as the evolving \LEGO{} structure itself can often block the line of sight, and the added bulk hampers reachability when assembling clutered \LEGO{} structures (shown in \cref{fig:overview-figure}(c)). These challenges highlight the need for localized, end-effector-level perception to ensure precise tool-to-brick alignment and robust error correction during dynamic assembly tasks.

\subsection{Contributions}
To address these challenges, this paper presents Eye-in Finger, a hardware-software co-design approach for accurate, occlusion-free \LEGO{} position estimation in robotic assembly. Specifically, our contributions include:
\begin{enumerate}
    \item A novel end-of-arm tool (EOAT) design with an embedded camera, integrating a vision system directly into the contacting tool tip. The design enables precise perception and guarantees a clear view of the object to manipulate.
    \item A vision-based pipeline capable of precisely estimating \LEGO{} brick poses and correct misalignment 
    \item Experimental validation demonstrating significantly improved accuracy and robustness over prior approaches.
\end{enumerate}
By integrating hardware innovations with real-time perception algorithms, this work provides a practical solution for vision-guided \LEGO{} manipulation and offers insights into addressing occlusion and precision challenges in robotic assembly.

\section{Related Work}
\subsection{Eye-in-hand Visual Servoing}

    Recent studies have successfully integrated cameras in grippers that use visual servoing to grasp items \cite{Khan2016,gupta2025svm}. Typically, these methods employ a camera mounted between the gripper to maintain an unobstructed view of the target object, thereby shrinking the footprint and reducing occlusion from the gripper itself. Their approaches have mainly targeted larger objects with high tolerances, such as harvesting tomatoes or grasping door handles.
   These configurations are less suitable for tasks that require a small stationary tool instead of grippers, such as \LEGO{} assembly, where the toolhead must remain fixed. Therefore, our approach integrates the camera inside the toolhead—further constraining its size. We also emphasize fine-grained local perception, achieving the sub-millimeter accuracy necessary for manipulating \LEGO{} bricks.
    
\subsection{Tactile Feedback}
    
    High-precision feedback has been achieved through vision-based tactile sensing, which demonstrates promise for accurately localizing small objects and performing insertion tasks with tight tolerance \cite{wong2012, 8202149}. However, these techniques usually require specially manufactured sensing hardware, which is generally costly.
    In addition, contemporary tactile sensors require specialized soft materials at the point of contact, which imposes additional challenges for applications such as \LEGO{} manipulation, which requires a rigid and firm contact to force snapping over the knobs on the bricks to establish stable connections. 
    Our approach, by contrast, achieves comparable perceptual accuracy with far lower cost and do not require physical contact, thereby broadening the range of applicable tasks.

\begin{figure*} 
    \centering
    \subfigure[Knobs segmentation.]{
        \includegraphics[width=0.17\textwidth]{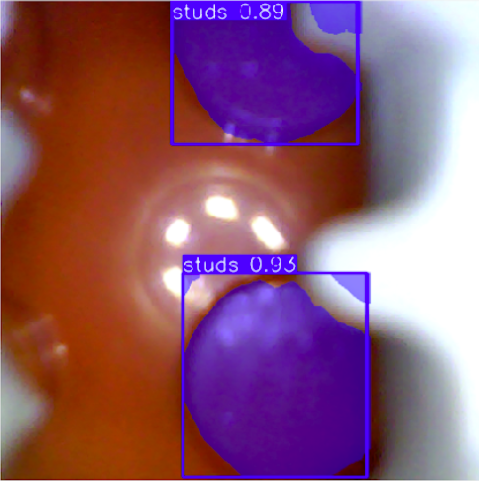}
        \label{fig:compute_offset_s1}
    }
    \hfill
    \subfigure[Masking.]{
        \includegraphics[width=0.17\textwidth]{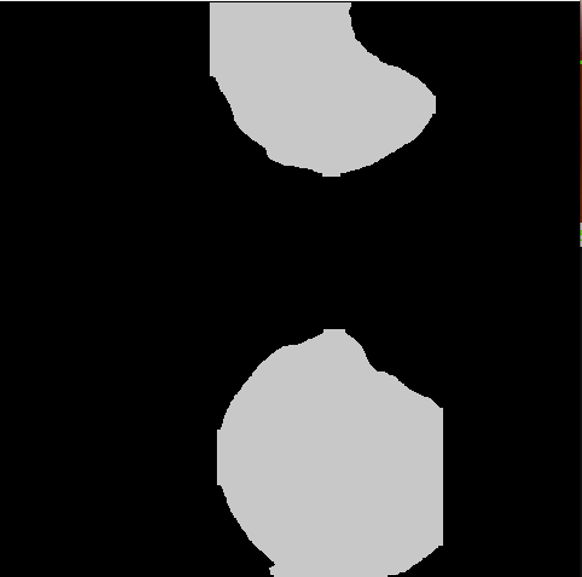}
        \label{fig:compute_offset_s2}
    }
    \hfill
    \subfigure[Bounds estimation.]{
        \includegraphics[width=0.17\textwidth]{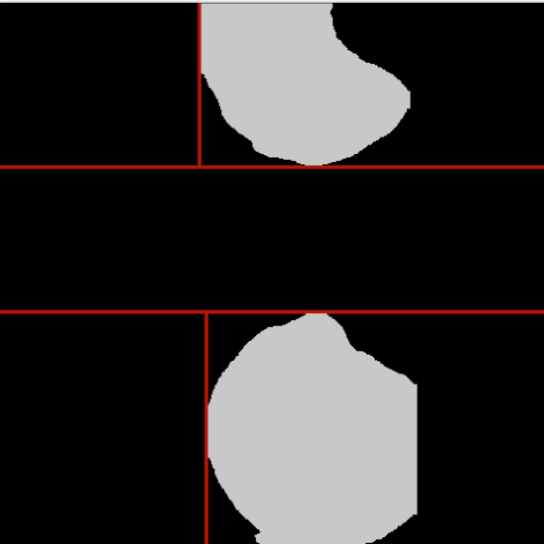}
        \label{fig:compute_offset_s3}
    }
    \hfill
    \subfigure[Silhouette reconstruction.]{
        \includegraphics[width=0.17\textwidth]{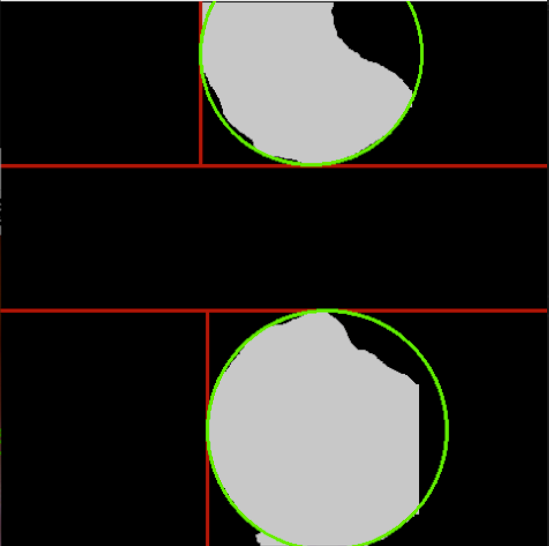}
        \label{fig:compute_offset_s4}
    }
    \hfill
    \subfigure[Compute offset.]{
        \includegraphics[width=0.17\textwidth]{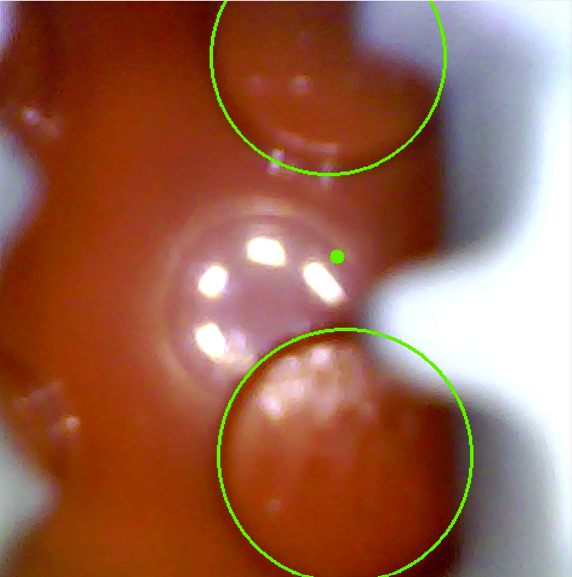}
        \label{fig:compute_offset_s5}
    }
    \hfill
    \caption{Illustration of the pipeline for pose estimation from partially occluded \LEGO{} target.}
    \label{fig:compute_offset}
    \vspace{-15pt}
\end{figure*}

\subsection{\LEGO{} Manipulation}
    
    Prior works \cite{popov2017data, fan2018learning} propose reinforcement learning based \LEGO{} manipulation policies but are difficult to transfer to the real world for customized \LEGO{} assembly. A recent study \cite{Liu2023} presents a hardware-software co-design that enables industrial robots to manipulate \LEGO{} bricks efficiently. This system leverages specialized EOAT designs and the inherent precision of industrial robots to perform \LEGO{} assembly tasks. \cite{huang2025apexmr} extends the EOAT to a bimanual system and enables robots to build customized and delicate \LEGO{} structures. However, these works operate with only force feedback, necessitating accurate calibration, and are susceptible to the accumulation of tolerance errors, particularly in complex, delicate assemblies. Our work addresses these challenges by incorporating real-time feedback mechanisms to correct positioning errors continuously during the assembly process.

\section{Methodology}

This paper aims to enhance both sensing and manipulation accuracy in robotic \LEGO{} assembly and disassembly. By integrating an endoscope camera directly into the tool, our Eye-in-Finger (EiF) design overcomes occlusion issues and improves perception at the end-effector level. This setup enables precise alignment, robust error correction, and real-time adjustments during assembly tasks. The following subsections detail A) the EiF hardware design, B) a novel pose estimation method for partially occluded targets, and C) a tilt estimation technique leveraging light reflections, all of which contribute to high-precision manipulation.

\subsection{Eye-in-Finger Hardware Design}
To address occlusion from other \LEGO{} structures and maintain a low-cost, compact profile that allows the robot to perform assembly tasks without additional reachability constraints, EiF is designed as shown in \cref{fig:overview-figure}(a). EiF’s mechanical functionality follows the design introduced in \cite{Liu2023,huang2025apexmr}, featuring an interface that securely snaps over knobs on a \LEGO{} brick. This ensures robust \LEGO{} manipulation while maintaining compatibility with various robotic platforms.

EiF integrates an endoscope camera, positioned directly behind the hollow section that interfaces with the \LEGO{} brick’s knobs as illustrated in \cref{fig:overview-figure}(a). This type of camera is selected because of its low cost, small form factor, and built-in LED to ensure consistent lighting. By aligning the interface and camera concentrically, we ensure continuous visibility of the target brick as long as it remains within EiF's reach. The camera is placed as far forward as possible within EiF to maximize the field of view while ensuring that the interface geometry does not obstruct the built-in LEDs. 


\subsection{Pose Estimation from Partially Occluded \LEGO{} Target}
\label{sec:3b}

To manipulate a \LEGO{} brick as illustrated in \cref{fig:overview-figure}, a rough pickup location is provided from the upstream planning module.
An image is captured from the EiF's integrated camera positioned directly above the estimated pickup location, with a height $z$ from the target \LEGO{} brick. 
Since the camera is integrated inside the EiF, visibility remains limited, leading to possible partial occlusions of the brick’s knobs as shown in \cref{fig:compute_offset_s1}. To address this, we estimate the knobs' positions based on their partially occluded contours, then compare them to the expected positions when the EiF is correctly aligned to compute the misalignment. The proposed solution is visualized in \cref{fig:compute_offset},  
The image from the camera is segmented using a fine-tuned YoloV8-seg model \cite{yolov8} as shown in \cref{fig:compute_offset_s1} and generate the corresponding mask as shown in \cref{fig:compute_offset_s2}. We designate the vertical pair of knobs that are closest to the center of EiF as intended target, and find the circles that best fit their masks. The process is summarized in \cref{alg:knob_center}.




In particular, for each mask, we define one horizontal bounding line $l_h$ and one vertical bounding line $l_v$ on its side closer to the center of Eif, as shown in \cref{fig:compute_offset_s3}. The reconstructed circle of the knob should be tangent to its bounding lines. The bounding lines are known to have high accuracy because the segmentation model performs well near the center of the image where shadows and occlusions are minimum. This is \cref{alg:knob_center} line 2-3.

\begin{algorithm}[b]
\caption{Estimate Center of Knobs}\label{alg:knob_center}
\begin{algorithmic}[1]
\For {each masks}:
    \State Identify bounding lines $l_h$, $l_v$
    \State Define a circle of radius $r$ tangent to both $l_h$ and $l_v$, on the side closer to the tool center
    \State Find $r$ that minimizes $C(r)$(Equation~\eqref{eq:cost_function})
    \State Compute knob center as the circle’s center
\EndFor
\end{algorithmic}
\end{algorithm}

After identifying these bounding lines, the next step is to determine the radius of the circles to accurately locate their centers. This is achieved through an optimization function, where the cost is defined as:

\begin{equation}
C(r) = \alpha \cdot \left(1 - \frac{|Mask \cap Circle|}{|Mask|} \right) + \beta \cdot \left|\mathbb{E}_{area}(z) - \pi r^2\right|
\label{eq:cost_function}
\end{equation}
The first term represents the percentage of the mask not enclosed by the circle, while the second term quantifies the difference between the circle's actual area and the expected area based on the estimated camera-to-brick distance $z$. 
$\alpha$ and $\beta$ weights to balance their influence, as the first term is a percentage (ranging from 0 to 1), and second term is size in pixels. The exact values are tuned empirically. This function enhances robustness against both z-axis inaccuracies and noisy masks. This is \cref{alg:knob_center} line 4-5, and produces the reconstructed circles shown in \cref{fig:compute_offset_s4}.

The center of the two knobs are then used to compute the offset between the planned and actual brick positions in \( x \), \( y \), and yaw, enabling visual servoing for precise misalignment correction.
Note that \LEGO{} bricks have identical knobs, which are indistinguishable from one another given only the local view. Thus, this method assumes that the initial estimated position, provided from the upstream module, is within half a knob’s distance from the true target position, or else the algorithm may lead the robot to move to a pair of nearby knobs that is closer.

\subsection{Tilt Estimation with Reflection}
\label{sec:3c}

\begin{figure}  
    \centering

    \subfigure[Tilted \LEGO{} brick]{
        \includegraphics[width=0.21\textwidth]{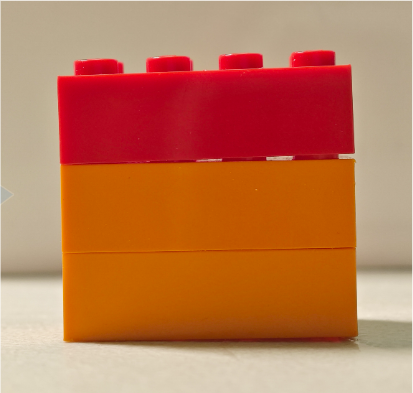}
        \label{fig:tilt1}
    }
    \hfill
    \subfigure[Reflection on \LEGO{} brick]{
        \includegraphics[width=0.195\textwidth]{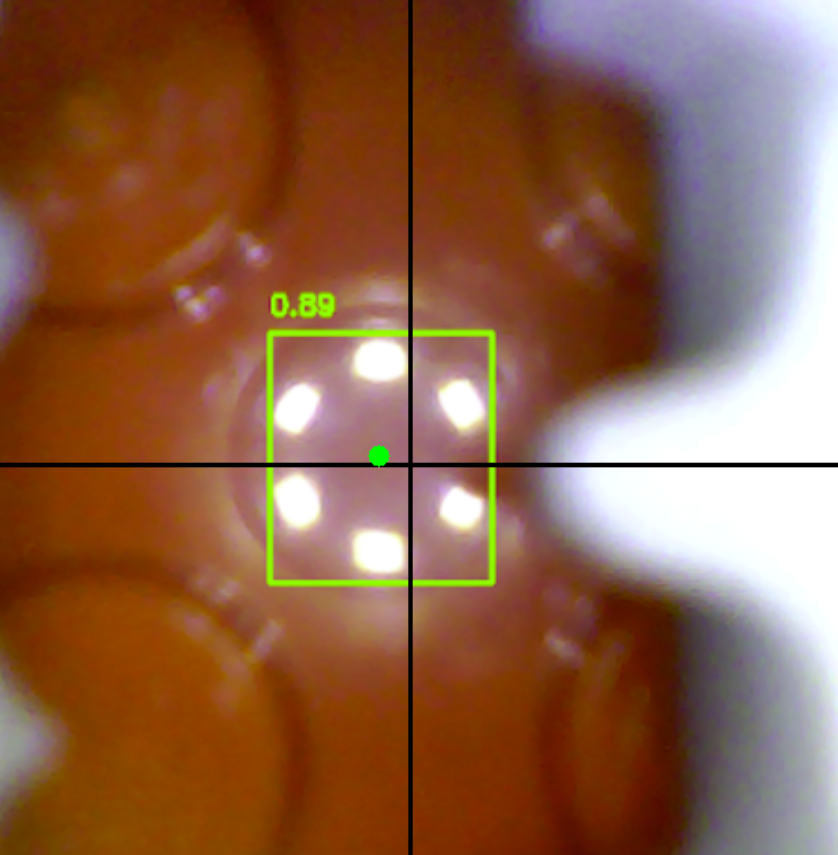}
        \label{fig:tilt2}
    }
    \hfill
    \subfigure[Measuring tilt with light reflection]{
        \includegraphics[width=0.350\textwidth]{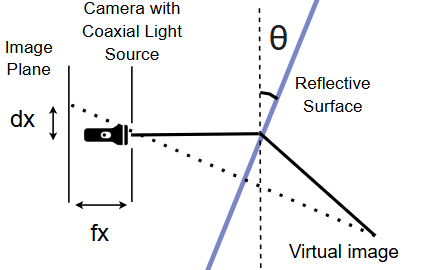}
        \label{fig:tilt3}
    }
    \caption{Illustration of tilt estimation with reflection. \label{fig:tilt}}
    \vspace{-15pt}
\end{figure}

The proposed method in \cref{sec:3b} enables robust recovery from planar misalignment errors, \ie on the XY plane. However, \LEGO{} assemblies can also have a slight tilt between bricks that are not firmly pressed together, as shown by \cref{fig:tilt1}, causing structural weakness.
To further improve the system robustness, we leverage the smooth and reflective surface of \LEGO{} bricks and detect tilt by analyzing the reflections of the built-in ring of LEDs in the endoscope camera, as seen in \cref{fig:tilt2}. 
Since the ring of light is concentric with the camera, its reflection appears at the center of the image if the brick is parallel to the image plane. As the brick tilts, the reflection shifts accordingly (\ie the green dot deviates from the black crossing point in \cref{fig:tilt2}), enabling the calculation of roll and pitch angles. \cref{fig:tilt3} illustrates the tilt detection. The tilt angle \( \theta \) is determined as
\begin{equation}\label{eq:tilt}
    \begin{bmatrix}
\theta_x \\
\theta_y
\end{bmatrix} = \begin{bmatrix}
\arctan\left(\frac{d_x}{f_x}\right) \\
\arctan\left(\frac{d_y}{f_y}\right)
\end{bmatrix}
\end{equation}
where $d_x, d_y$ are the light source's position (in pixels) away from center, and $f_x, f_y$ are the focal lengths in pixels of the camera.
This approach provides highly precise measurements when \( \theta \) is near 0°. For instance, with a 480p camera setup and \( (f_x, f_y) = (830, 830) \), the theoretical accuracy is 0.07° per pixel near 0° tilt, though actual accuracy is affected by detection noise. The setup has a theoretical range of ±16° in each direction before the LED ring center becomes unobservable.

We use YOLO V11 \cite{yolo11}, which is the state-of-the-art model for real-time bounding box detection, to perform tilt detection. Specifically, we fine-tune YOLO V11 on collected images with reflections.
Since the model assumes that light reflects off the flat \LEGO{} brick surface rather than the knobs, visual servoing is used to align the expected light reflection center to maximize measurement range. For 1-knob-wide bricks, this target location is the midpoint between two adjacent knobs. For bricks with width $\geq$ 2 knobs, it is the center between four adjacent knobs.

\subsection{\LEGO{} Manipulation Process}



\begin{algorithm}[t]
\caption{\LEGO{} Manipulation Process \label{alg: manipulation}}
\begin{algorithmic}[1]
\State \textbf{Step 1: Peek}
\State Move to placing position
\While{tool is not centered}
    \State Move towards center
\EndWhile
\State Record position

\State \textbf{Step 2: Pick Up}
\State Move to pick-up position
\While{tool is not centered}
    \State Move towards center
\EndWhile
\State Pick up brick

\State \textbf{Step 3: Place}
\State Move to recorded position
\State Place brick at recorded position
\State \textbf{Step 4: Inspection}
\State record brick position (using \textbf{Peek})
\State record brick tilt

\end{algorithmic}
\end{algorithm}

Our Eye-in-Hand solution is applied to \LEGO{} manipulation, as illustrated in \cref{alg: manipulation}. When picking up a \LEGO{} brick, the robot initially moves to an estimated brick position provided by the upstream planner. Due to the minimal visual differences between \LEGO{} knobs, the visual servoing pipeline assumes that the two knobs closest to the center of the gripper are the intended target. The algorithms described in \cref{sec:3b} and \cref{sec:3c} compute the positional and orientation offsets between the current position and the target, and the robot adjusts its position accordingly. This process is repeated until the offset falls below a predefined threshold, at which point the robot proceeds to pick up the brick. This corresponds to \cref{alg: manipulation} line 8-12

When placing a \LEGO{} brick, the tool head is obstructed by the brick it is holding, preventing direct visual inspection of the placement site. To address this, the placement site is inspected before picking up a brick. The same process is used to compute the offset between the actual and estimated placement positions, which is recorded and later compensated during placement. This is \cref{alg: manipulation} line 2-6. Since industrial robots exhibit high precision, any assembly errors can only be introduced during the placement step; there is negligible difference between this pre-measured offset and that during subsequent placements.

When necessary, the program performs an inspection step after placing the brick. This is \cref{alg: manipulation} line 16-18.This includes a position inspection similar to that conducted during placement, as well as pitch and roll measurements described in \cref{sec:3c}. This process enables the automatic detection of loosely placed bricks or other assembly defects.


\section{Experiments}

To demonstrate the effectiveness of the proposed EiF in \LEGO{} manipulation, we conduct the following experiments:
\begin{enumerate}
    \item Compare measurement accuracy of \LEGO{} bricks in X, Y, Yaw, Pitch, Roll against wrist mounted depth camera
    \item Perform pick (\ie disassemble) and place (\ie assemble) of \LEGO{} bricks with noise-injected calibration in order to verify its robustness against calibration and tolerance errors
    \item Evaluate efficiency and accuracy of human teleoperation of the robot, with and without the aid of EiF
\end{enumerate}
Note that we do not consider the offset in the Z direction since it can be addressed using direct force feedback \cite{huang2025apexmr}. To remove any fabrication inaccuracy, we calibrate the system by manually align EiF with a \LEGO{} brick, and set the expected center of knobs (in \cref{sec:3b}) and reflection (in \cref{sec:3c}) base on that reference reading. Coefficient $\alpha$ and $\beta$ in \cref{eq:cost_function} are $3.5 \times 10^4$ and $ 1 $ respectively.

\subsection{Measurement Accuracy}

To compare the precision of our EiF approach with common wrist-mounted depth camera setups such as \cite{kang2021, black2024pi_0, lan2025bfa}, we attach an Intel RealSense D405 camera to the wrist link of a Yaskawa GP4 robot, positioned 7 cm away from a \LEGO{} brick—the minimum ideal range specified by the manufacturer. A 4$\times$2 \LEGO{} brick was placed on a stationary board, starting with the brick centered in the camera’s field of view. 

To evaluate position accuracy along the X and Y axes, we incrementally jog the robot in 0.5 mm steps for three iterations in each direction, resulting in a total of 49 measurements. The relative pose between the camera and the \LEGO{} brick was recorded and compared against the ground truth position of the robot. 
For the yaw measurement, the robot was rotated in 2-degree increments for four iterations in each direction, leading to nine total measurements. Position measurements were obtained using the YoloV8-seg \cite{yolov8} segmentation model.

To assess roll (rotation along the long axis) and pitch (rotation along the short axis) measurements, the camera was initially aligned parallel to the brick. The pitch and roll angles were then incremented by 2 degrees for three iterations in each direction, totaling 49 measurements. These measurements were derived by comparing depth values at both ends of the block. The same process was repeated for our EiF approach, utilizing the method described in \cref{sec:3b} for x, y, and yaw calculations, and the approach in \cref{sec:3c} for roll and pitch estimation.

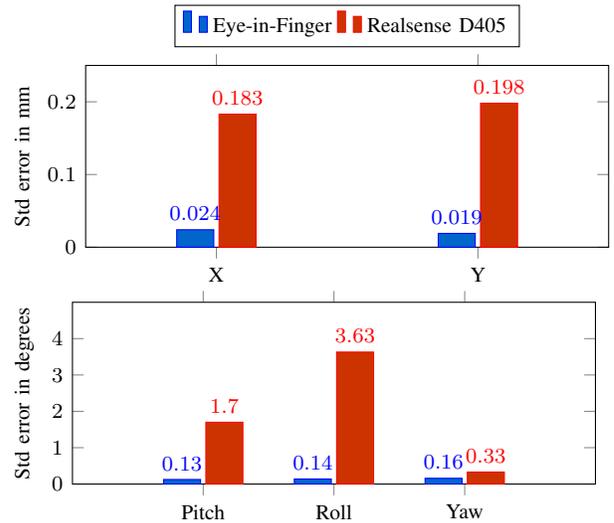
\begin{figure}  
    \centering
    \parbox{0.48\textwidth}{
        \begin{tikzpicture}
    \begin{axis}[
        ybar,
        ymin=0, ymax=0.25,
        ytick={0,0.1,0.2},
        symbolic x coords={X,Y},
        xtick=data,
        ylabel={Std error in mm},
        ylabel near ticks,
        legend style={at={(0.5,1.1)},anchor=south,legend columns=-1},
        nodes near coords,
        every node near coord/.append style={
            /pgf/number format/fixed,
            /pgf/number format/precision=3
        },
        enlarge x limits=0.5,
        bar width=14pt,
        height=4cm,
        width=\linewidth,
        font=\footnotesize,
        scaled y ticks=false, 
        yticklabel style={/pgf/number format/fixed} 
    ]
    \addplot+[ybar, fill=EyeInFingerColor] coordinates {(X,0.024) (Y,0.019)};
    \addplot+[ybar, fill=RealsenseColor] coordinates {(X,0.183) (Y,0.198)};
    \legend{Eye-in-Finger, Realsense D405}
    \end{axis}
\end{tikzpicture}
    }
    \parbox{0.48\textwidth}{
        \begin{tikzpicture}
    \begin{axis}[
        ybar,
        ymin=0, ymax=5,
        ytick={0,1,2,3,4},
        symbolic x coords={Pitch, Roll, Yaw},
        xtick=data,
        ylabel={Std error in degrees},
        ylabel near ticks,
        legend style={at={(0.5,1.1)},anchor=south,legend columns=-1},
        nodes near coords,
        every node near coord/.append style={
            /pgf/number format/fixed,
            /pgf/number format/precision=2
        },
        enlarge x limits=0.5,
        height=4cm,
        bar width=14pt,
        width=\linewidth,
        font=\footnotesize,
        scaled y ticks=false, 
        yticklabel style={/pgf/number format/fixed} 
    ]
    \addplot+[ybar, fill=EyeInFingerColor] coordinates {(Pitch,0.125) (Roll,0.139) (Yaw,0.158)};
    \addplot+[ybar, fill=RealsenseColor] coordinates {(Pitch,1.70) (Roll,3.63) (Yaw,0.33)};
    \end{axis}
\end{tikzpicture}
    }
    \caption{Measurement Accuracy \label{fig: sd_error}}
\end{figure}

As shown in \cref{fig: sd_error}, our approach delivers an 8.8-fold improvement in position accuracy, 1.9-fold in yaw, 13.6-fold in pitch, and 26.1-fold in roll. The improvement is particularly significant for roll accuracy, where small tilts along the long axis produce minimal depth differences, making it difficult to perceive using depth sensing.

\subsection{Calibration Error Correction}

To test the effect of EiF in improving the robustness of \LEGO{} manipulation, we test the success rate of picking and placing of \LEGO{} bricks in \textit{inaccurately} calibrated conditions. After calibrating to a ground truth position, we introduced artificial calibration errors of magnitude \( \delta \) with random orientations. The robot then attempted to pick up or place a \LEGO{} brick on the board while compensating for these errors. The experiment was repeated 12 times for each error magnitude \( \delta \).
\begin{figure}  
    \centering
    \begin{tikzpicture}
    \begin{axis}[
        xlabel={Calibration Error (mm)},
        ylabel={Pickup Success Rate},
        ymin=0, ymax=1.2,
        xmin=0, xmax=3.2,
        ytick={0,0.2,0.4,0.6,0.8,1.0},
        xtick={0,0.4,0.8,1.2,1.6,2.0,2.4,2.8,3.2},
        legend style={at={(0.5,1.1)},anchor=south,legend columns=-1},
        height=4cm,
        width=\linewidth,
        font=\footnotesize
    ]
    \addplot+[thick, mark=., RealsenseColor] coordinates {(0,1) (0.4,1) (0.8,0.83) (1.2,0.16) (1.6,0.06) (2.0,0) (2.4,0) (2.8,0) (3.2,0)};
    \addplot+[thick, mark=., EyeInFingerColor] coordinates {(0,1) (0.4,1) (0.8,1) (1.2,1) (1.6,1) (2.0,1) (2.4,0.75) (2.8,0.61) (3.2,0.47)};
    \legend{Open Loop, Eye-in-Finger}
    \end{axis}
\end{tikzpicture}
    \vspace{-30pt}
    \begin{tikzpicture}
    \begin{axis}[
        xlabel={Calibration Error (mm)},
        ylabel={Place Success Rate},
        ymin=0, ymax=1.2,
        xmin=0, xmax=3.2,
        ytick={0,0.2,0.4,0.6,0.8,1.0},
        xtick={0,0.4,0.8,1.2,1.6,2.0,2.4,2.8,3.2},
        legend style={at={(0.5,1.1)},anchor=south,legend columns=-1},
        height=4cm,
        width=\linewidth,
        font=\footnotesize
    ]
    \addplot+[thick, mark=., RealsenseColor] coordinates {(0,1) (0.4,1) (0.8,0.83) (1.2,0.5) (1.6,0.05) (2.0,0) (2.4,0) (2.8,0) (3.2,0)};
    \addplot+[thick, mark=., EyeInFingerColor] coordinates {(0,1) (0.4,1) (0.8,1) (1.2,1) (1.6,1) (2.0,1) (2.4,0.89) (2.8,0.67) (3.2,0.55)};
    \end{axis}
\end{tikzpicture}
    \vspace{-30pt}
    \caption{Comparison of the success rates between Lego manipulation with and without Eye-in-Finger under different magnitudes of calibration errors. 12 trials were conducted for each method at each magnitude}
    \label{fig: pick and place success}
    \vspace{-15pt}
\end{figure}
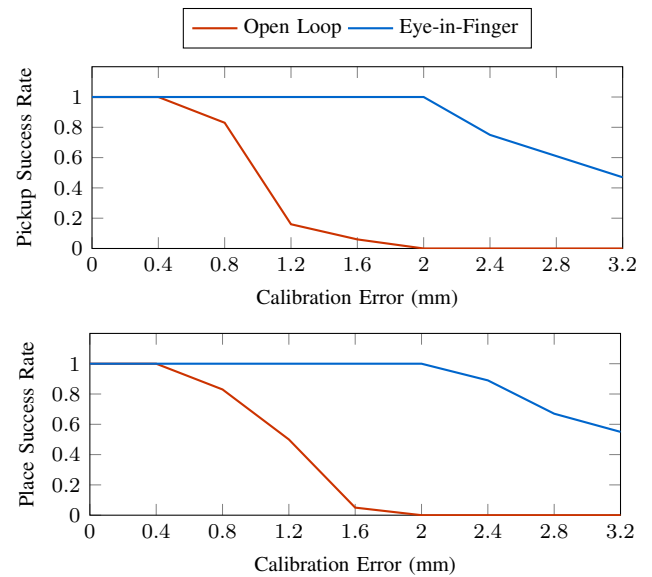

\Cref{fig: pick and place success} compares \LEGO{} manipulation performance under calibration errors, using both a baseline open-loop system and the Eye-in-Finger visual feedback. Without visual feedback, the robot successfully manipulated bricks for small errors, aided by the tool’s chamfered design that facilitates self-centering. However, success rates dropped sharply beyond a 0.8 mm error. In contrast, with visual feedback from Eye-in-Finger, success remained at 100\% for errors up to 2.0 mm. With higher error however, the segmentation model becomes less reliable due to increased occlusion, causing success rate to drop. These results highlight the effectiveness of real-time visual feedback in reducing dependency on precise calibration and mechanical accuracy for \LEGO{} manipulation.

\subsection{Eye-in-Finger Aided Teleoperation}
\begin{figure}  
    \centering
    \includegraphics[width=0.8\linewidth]{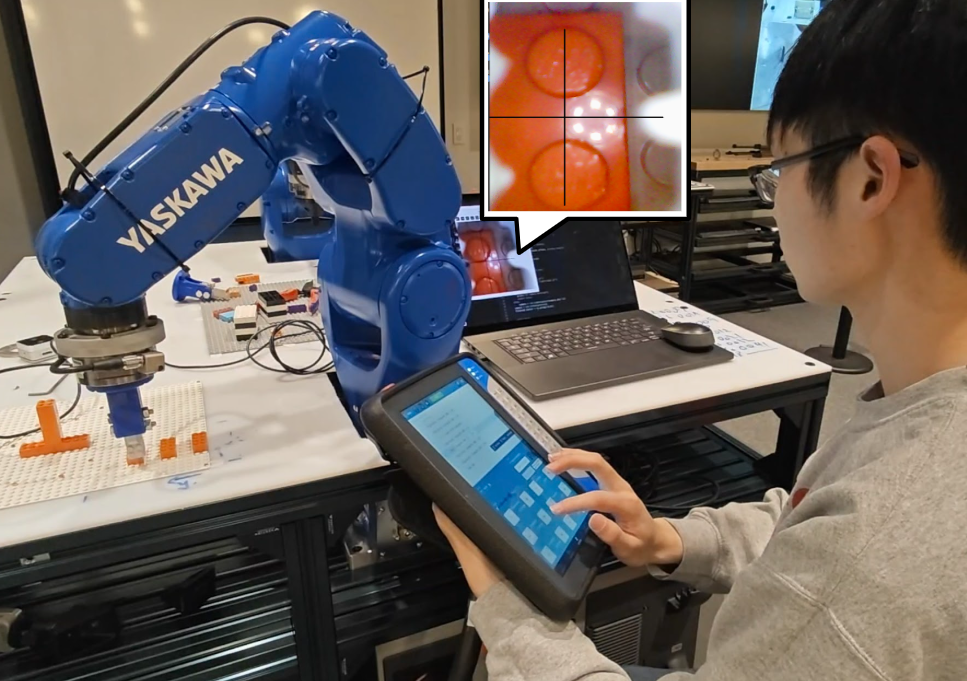}
    \caption{EiF view aided teleoperation \label{fig:teleop}}
    \vspace{-15pt}
\end{figure}
Beyond enabling precise closed-loop control, Eye-in-Finger also enhances human teleoperation. In \LEGO{} manipulation, accurately aligning the EOAT with a brick is a slow and non-trivial task, requiring the operator to visually inspect alignment from multiple angles. With perception embedded within the EOAT, as shown in \cref{fig:teleop}, the operator can rely on the live top-down camera view for fine alignment, making the process more intuitive and efficient.

\begin{figure}  
    \centering
    \includegraphics[width=0.35\textwidth]{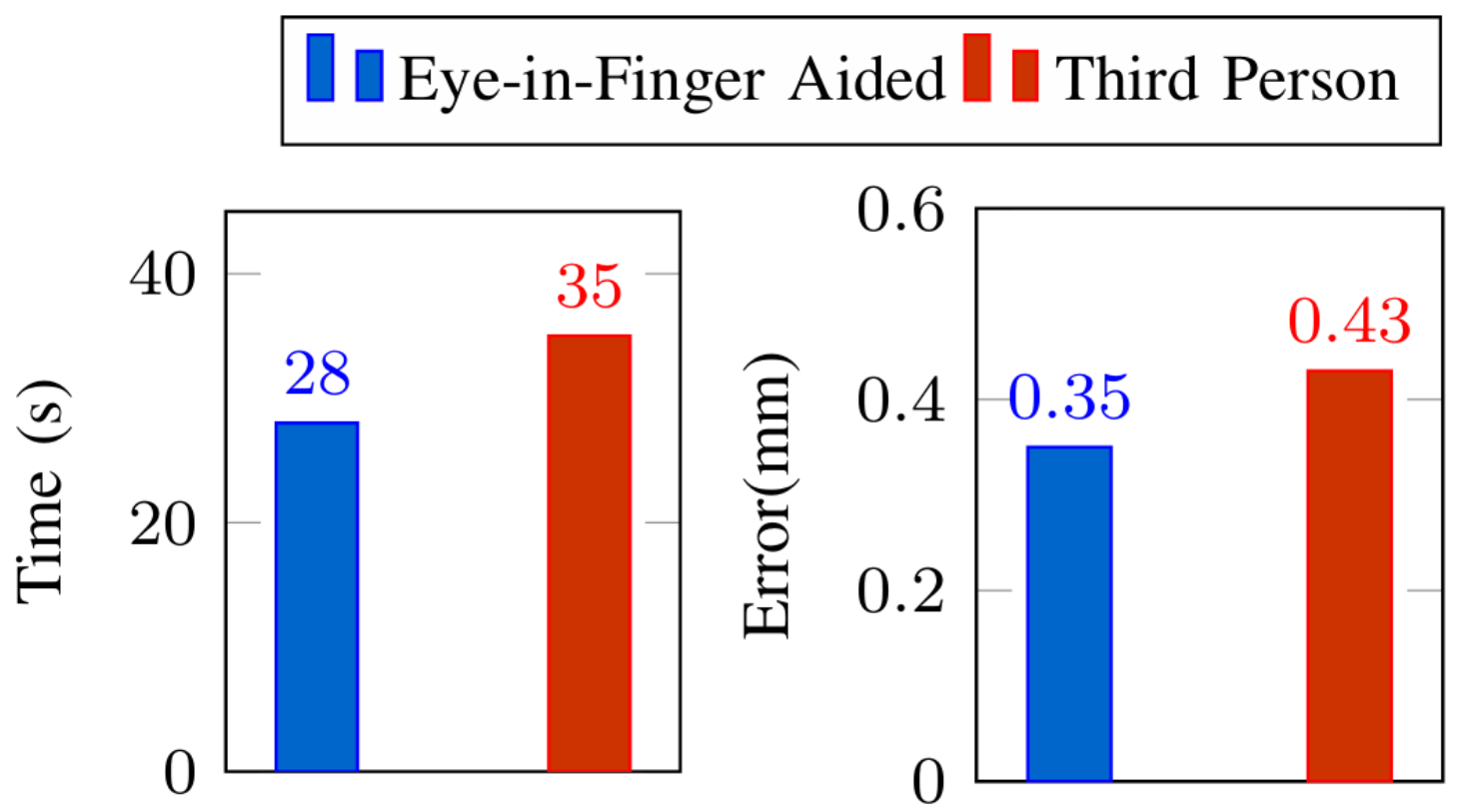}
    \caption{Comparison between time and accuracy of human tele-operation with only third person view or with the additional information provided by EiF.}
    \label{fig: teleoperation comparison}
    \vspace{-15pt}
\end{figure}

\Cref{fig: teleoperation comparison} compares human teleoperation efficiency and accuracy using a traditional third-person view versus the Eye-in-Finger system. In the experiment, a \LEGO{} brick was randomly placed on a stationary board, approximately 10 cm from the EOAT in a random direction. The human operator was tasked with picking up the brick, using either a third-person view or the live stream from the embedded camera. Task completion time and pickup accuracy were recorded. 

Based on 10 human trials, operators using the Eye-in-Finger system completed the task 17\% faster and achieved a 19\% reduction in positioning error compared to those relying solely on a third-person view. This demonstrates the benefits of real-time perception within the EOAT for improving human-robot interaction and operational efficiency.

\section{Discussion}
In this paper, we presented Eye-in-Finger, a novel hardware-software co-design approach that enables perception at the point of manipulation. This design mitigates occlusion issues and provides precise visual feedback, enhancing the accuracy of fine manipulation tasks. We evaluated its effectiveness in the context of \LEGO{} manipulation. Experimental results demonstrate that our proposed approach significantly outperforms wrist-mounted depth cameras in measurement accuracy. Additionally, our method substantially improves the robustness of manipulation by leveraging real-time visual feedback and robust error detection. 

\section{Limitations and Future Works}
Despite the remarkable improvements Eye-in-Finger introduces in \LEGO{} manipulation, several factors limit its applicability. Most importantly, the limited field of view of the camera inside EiF necessitates an estimated position of the target object. In the \LEGO{} manipulation scenario, this estimation is achieved by calibrating the build plate and specifying the initial positions of all \LEGO{} bricks. For broader applications, an additional camera would be required to provide high-level perception, guiding the robot towards an estimated target position.

In the future, we aim to enhance Eye-in-Finger by integrating it with additional a global camera to overcome the field-of-view constraint and enable more comprehensive perception. Additionally, we aim to incorporate learning-based manipulation policies to improve dexterity, allowing the system to handle a wider range of tasks beyond LEGO assembly. This integration would enable more dexterous and intelligent manipulation in complex, unstructured environments.

\bibliographystyle{IEEEtran}
\bibliography{reference}     

\end{document}